# 영역 기반 Deep CNN 을 사용한 농지에서의 잡초 식별

# Farm land weed detection with region-based deep convolutional neural networks

○모하마드 이브라힘 싸커*, 김형석**
Mohammad Ibrahim sarker*, Hyongsuk Kim**
* 전북대학교 전자공학과 (TEL: 063-270-2477; E-mail: sarkeribrahim@gmail.com)
** 전북대학교 전자공학과, 지능형로봇연구센터 (TEL: 063-270-2477; E-mail: hskim@jbnu.com)

**Abstract** Machine learning has become a major field of research in order to handle more and more complex image detection problems. Among the existing state-of-the-art CNN models, in this paper a region-based, fully convolutional network, for fast and accurate object detection has been proposed based on the experimental results. Among the region based networks, ResNet is regarded as the most recent CNN architecture which has obtained the best results at ImageNet Large-Scale Visual Recognition Challenge (ILSVRC) in 2015. Deep residual networks (ResNets) can make the training process faster and attain more accuracy compared to their equivalent conventional neural networks. Being motivated with such unique attributes of ResNet, this paper evaluates the performance of fine-tuned ResNet for object classification of our weeds dataset. The dataset of farm land weeds detection is insufficient to train such deep CNN models. To overcome this shortcoming, we perform dropout techniques along with deep residual network for reducing over-fitting problem as well as applying data augmentation with the proposed ResNet to achieve a significant outperforming result from our weeds dataset. We achieved better object detection performance with Region-based Fully Convolutional Networks (R-FCN) technique which is latched with our proposed ResNet-101.

## I. Introduction

In recent years, as the world population has grown, existing land and natural resources have decreased more than ever, leading to precision agriculture increasingly capturing more attention of the researchers. In agriculture, weeds control is a time-consuming and expensive activity. Moreover, the long-term use of herbicide is a potential source of pollution, which could damage people, animals and the environment. In fact, agricultural herbicides have been regularly sprayed in fields and overused in a conventional way for several years. Manual spraying has caused severe environmental pollution. Therefore, the research efforts are being encouraged for designing weeds-detecting technologies for precision spraying of selective herbicides with the final purpose of saving herbicides and reducing environmental pollution without sacrificing the crop yield. One alternative to efficiently applying herbicides consists of using an Artificial intelligence system to detect weeds autonomously and precisely. Locating and identifying multiple items in an image is something that is still difficult for machines to accomplish.

However, significant progress has been made in the last few years on object detection with convolutional neural networks (CNNs). The problem of shallow vs deep neural networks has been in discussion for a long time in machine learning [1] and has shown that shallow networks require exponentially more components than deeper networks.

Nowadays the network for object detection is mostly divided into two parts, first they classify the object and then they make the boundary box to make the object classified. To classify the object researchers have used convolutional network with MLP and boundary box regression. Several other models can also be used such as R-CNN [8], Fast R-CNN [10], Faster R-CNN [11], R-FCN [7].

However, the state of art network for image classification such as ResNet [2] and GoogLeNet [3,4] use fully convolutional network. Based on this, we decided to use fully convolution network in weeds detection, but all attempts failed because of poor accuracy. The paper titled "Deep Residual Learning for Image Recognition [2]", adds RoI pooling layer, which improves accuracy, but it slows down speed. The problem occurs, as the network doesn't share weights when computing each RoI. There is a contradiction between translation-invariance in classification of images and translation-variance in processing object detection. Deep convolutional network is a beneficial network as it identifies the highly-deviated input images for classification purpose. In contrast, object's orientation needs translation-variance, such that it must generate the appropriate reasonable mapping relationship with candidate box when the object changes. We predict that the deeper convolutional networks are less sensitive to translation. To balance them, we use a Region-based Full Convolutional Network. Also in this network, there is a set of position sensitive score maps as fully convolutional layer's output, which includes the location information of the object. The front layer is RoI pooling layer which deals with space information, after it, without any weight layer. This method can solve the problem between translation-invariance and translation-variance. Also, we inserted dropout [5] after first pooling layer for successfully reducing overfitting problem, though previously dropout in residual network was studied in [6] where dropout was being inserted in the identity part block, and the author showed negative effect of that structure. Experimental results with dropout (at a ratio of 0.5) performed higher accuracy in our weeds dataset. Finally, we proposed our modified ResNet as deep CNN model with Region-based Fully Convolutional Network (R-FCN) [7] framework for weeds detection as well as for increasing object identification and accuracy.

## II. Proposed Method and Architecture

Large-scale ConvNets are not only successful in image classification tasks, but also in object detection tasks. Because ConvNets excel at classification of an image, they have been combined with image to satisfying our requirements we choose a particular network and can train the network with our own datasets. However, if the dataset of the particular detection problem is not big enough than it is possible that the deep CNN might be insufficiently trained. For resolving such types of problem, we can choose the trained deep CNN model and can perform the structural modification and then finetune the modified network with our own dataset to exhibit the desired results. Following R-CNN [8] we followed the most popular two-stage object detection strategy [8,9,10,11,12,13] which contains (i) region proposal, and (ii) region classification [14,15]. Region based detection system still leading on several benchmarks [16,17,18]. In our experiment, we tried following things to improve ResNet on our own dataset: The original paper used a "Building Block" shown in the left part of the Figure below (Fig:1(a)), inside the block there is Convolution Layer, Batch Normalization [19] and ReLu [20], followed by another Convolution Layer and Batch Normalization. We investigated few alternate strategies.

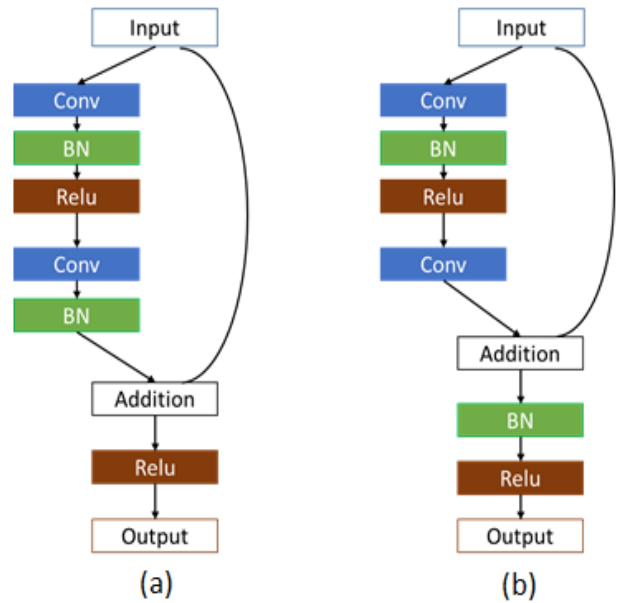

Figure 1: (a) Original model, and (b) Represents first alternate model.

If original model is compared with the first alternate model, then we can find that in Fig. 1 (b) batch normalization is relocated after the Addition. The reason behind this choice is to test whether normalizing the first term of the addition is desirable or not. It grew out of the mistaken belief that batch normalization always normalizes to have zero mean and unit variance. If this were true, building an identity building block would be impossible because the input to the addition always has unit variance. However, this is not true. BN layers have additional learnable scale and bias parameters, so the input to the batch normalization layer is not forced to have unit variance.

Removing the ReLu from second block is the second alternative strategy (Fig: 2 (b)). Noticing that in the reference architecture, the input cannot proceed to the output without being modified by a ReLu was the main idea behind this strategy. This makes identity connections technically impossible because negative numbers would always be clipped as they passed through the skip layers of the network. We could either move the ReLu before the addition or remove it completely to avoid this. However, moving the ReLu before the addition is not correct as such architecture would ensure that the output would never decrease because the first addition term could never be negative. While on other hand removing the ReLu completely i.e. simply sacrificing the nonlinear property of this layer is another option. In the test performance, improvement was observed in both the strategies. But since the improvements were of same intensity for both the strategy, it is difficult to decide which strategy is better.

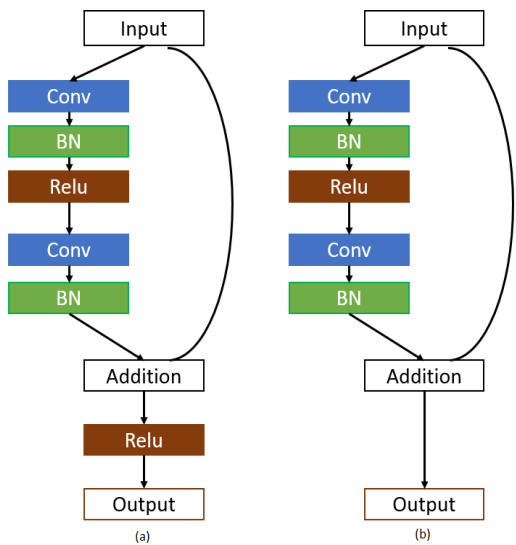

Figure 1: (a) represents the reference paper model, and (b) represents second alternate model.

We inserted RPN layers after conv and added one additional conv layer for size normalization. Normalization allows us to use much higher learning rates and be less critical about initialization. It also acts as a regularizer i.e. in some cases eliminating the need for dropout. We also made use of batch normalization after each of the convolutional layers to greatly speed up training.

Next, we tried adding one more convolutional layer just after Input layer but no improvements were observed, unlike the belief that deep learning's network performance should increase if we add more layers to it. Then we tried changing the hyper parameters, i.e. the learning rate (as we are using gradient decent for finding minimum error), the weight decay and the momentum. We changed momentum from 0.9 to 0.7. As we use gradient decent for updating weight, due to which weights will have some velocity, and due to inertia, the weights will try to do go back to old position, so we tried changing the momentum (by giving some velocity to our weights). There were some improvements on 100 iterations but after increasing the iteration the accuracy decreased, which was probably due to over fitting the data.

So finally, it became that in our network there was an overfitting [21] problem as we were using very deep network such as ResNet and using very small dataset to train it. This implies that indirectly we were forcing our network to learn by giving it more bias which eventually leads to overfitting. In our experiment, we used ResNet 101 with a small dataset (2000 images). Using such small dataset in very deep network can possibly cause overfitting as well as an inability to identify all objects. Thus, we should firstly add dropout in every residual block

In another model, we tried to use dropout after first Pooling layer (Figure: 3(a)) because as pooling itself reduces the dimension it gives less input data point as desired. Although we can even add dropout before Pooling but it will increase the computation since we will need to learn more weights which will also eventually lead to overfitting, as we have less input data. So, we added dropout layer because we wanted our system to be trained in such a manner that it should not learn our input data as we have very small dataset. Thus, in order to handle that problem, we added a dropout layer before the first residual block and checked whether the performance and efficiency of our proposed ResNet-101 increased or not. We found out that the addition of dropout before first residual block significantly enhanced the performance of the ResNet-101 for our dataset which we have discussed in experiment section.

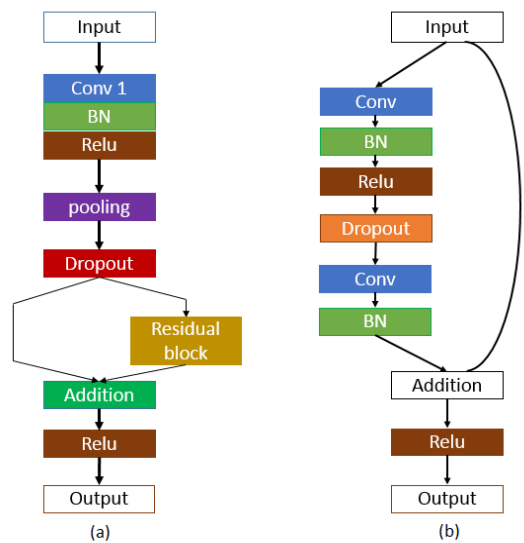

Figure 2: (a) Dropout after pooling, and (b) Dropout inside residual block.

Now In order to introduce the translation variance into the fully convolutional network, a special convolution layer is designed as the output of the fully convolutional network. The convolution layer outputs the position-sensitive score map, and each score map introduces the position Information, such as the top of the object. In the last layer of the network, there is a position-sensitive RoI pooling layer, to complete the detection of objects. In the whole network framework, all the layers that can be learned are convolution layers, and the spatial position information is introduced into the feature learning, so that the whole network can carry out end-to-end learning. The Proposed model for our weeds detection is showed in Figure: 4.

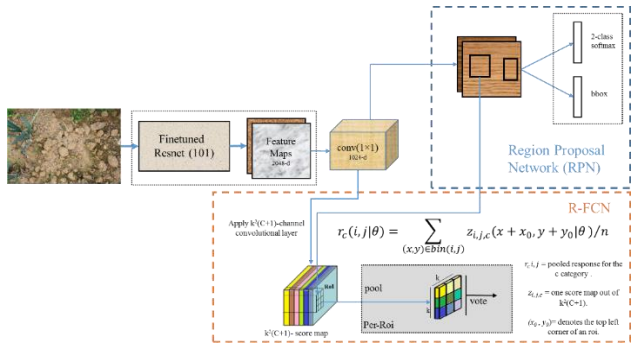

Figure 3: Proposed model for Farm land weed detection.

We use our modified ResNet-101 as the basis of the convolution network. ResNet-101's last convolutional feature map has a dimension of 2048, with a new convolution layer, which reduces the dimension to 1024 dimensions, so that the shared convolution layer is 101 layers and it generate $k^2(C+1)$ of the positive-sensitive score map of the channel. To generate RPNs proposal we use faster R-CNN. After a given proposal region (RoIs), the RPNs classify the RoIs as the target object or background. All the parameters can be shared in the R-FCN network, instead of fully connected layer, we use convolutional layer. At the end layer, we add one more layer which generates score maps and the total score maps are $k^2(c+1)$. All these score maps can store the feature map's spatial information i.e. position-sensitive score map which is a total of $k^2(c+1)$ channel output. This one $k^2$ score map corresponds to the spatial information describing a grid of k*k. For example, k*k=3*3, then the nine score maps are corresponding to the {top-left, top-center, top-right, …, bottom right} positions of the target classification. The last layer of R-FCN is a position sensitive RoI pooling layer, which generates value for each RoI from k*k for class classification. Figure 5 visualizes the positive sensitive RoI pooling layer.

Finally, SoftMax is used to determine the RoI class. At the same time $4^2k$ dimensional vector is generated for regression of the Boundary box which is similar to Faster R-CNN. This improved version of Faster RCNN uses same loss function as previous. Also, RFCN and RPN share the same network parameters, training methods and Faster R-CNN training strategies and procedures are basically the same. So, This R-FCN is called an improved version of Faster R-CNN.

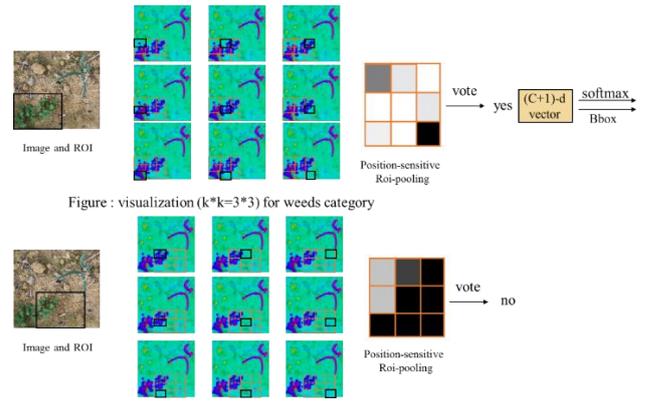

Figure 4: Visualization of position sensitive RoI pooling.

## III. Experimental Results

We performed experiments on our farmland's weeds dataset that has 3 object categories. In all our experiments, total training and validation data is 2000 and test data is 250. Object detection accuracy was measured by mean Average Precision (mAP). In this experiment, to find an optimal performing model, we did comparison with different CNN models using Faster R-CNN and R-FCN based on our own dataset. Moreover, we also made comparison with different proposed architectures to find the finest architecture. Furthermore, we compared our proposed architecture with the conventional CNN models using our own dataset.

Comparisons between different Deep CNN models using Faster RCNN: We used different Deep CNN models with standard Faster RCNN, results of which are shown in Table 1.

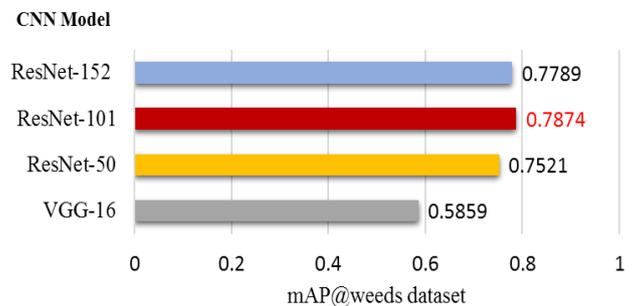

Table 1: Weed detection using Faster R-CNN and different Deep CNN models.

Comparisons between different Deep CNN models using Faster R-FCN: Next we used different Deep CNN models with R-FCN, results of which are shown in table 2. Similar to other research articles we also found that for our dataset, the accuracy of R-FCN is higher when compared to standard faster R-CNN.

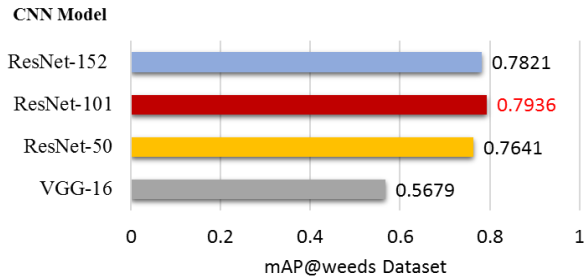

Table 2: Weed detection using R-FCN and different Deep CNN models.

It can be observed from Table 1 and Table 2 that the R-FCN with ResNet -101 achieves comparatively good accuracy than the faster RCNN with ResNet -101, which emphasis that we can get a high performance of desired results using R-FCN. As the probability of getting high performance results is better with R-FCN, so we finalized the ResNet -101 along with R-FCN as our experimental model and projected this experimental model with modification as a proposed model.

Next, we removed position-sensitivity by setting k=1 from R-FCN. This is equivalent to global pooling within each RoI. But this time we got mAP = 0.70, which is much lower than native Faster R-CNN. Later we tried different RoI output size on R-FCN which is shown in Table 3. It can be observed from Table 3 that the finest mAP is achievable with ROI output size as (7*7) for our weeds dataset.

| Method | ROI output size | mAP |
|---|---|---|
| R-FCN | 3*3 | 0.761 |
| R-FCN | 7*7 | 0.7701 |
| R-FCN | 9*9 | 0.7505 |

Table 3 Comparison with different RoI output size.

Further on, we tried to improve our accuracy by doing fine-tuning on ResNet -101. In chapter 4 we have already explained all alternative strategies of ResNet -101 and Table 4 shows the results of all explained strategies. For this experimental case, we have used ResNet -101 with R-FCN, and the RoI output size of R-FCN was 7*7. It is to be noticed in Table 4 that the modified ResNet -101 (by adding dropout after pooling layer) network significantly outperforms than the other strategies and thus we projected this modified ResNet -101 as proposed architecture for CNN model. Table 4 shows the results of alternative strategies which were applied on ResNet.

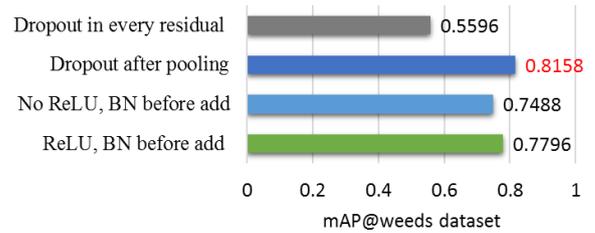

Table 4 Comparison with different alternative strategies in residual block.

The final results are shown in Table 5. Our proposed method got 81% accuracy. This is comparable to the Faster R-CNN baseline as well as R-FCN baseline. Our proposed method performs better in small dataset and also improves the object identification accuracy.

| Architecture | Dataset | mAP |
|---|---|---|
| Faster R-CNN (ResNet-101) | Weeds & Onion | 0.7874 |
| R-FCN (ResNet-101) | Weeds & Onion | 0.7936 |
| Proposed Model | Weeds & Onion | 0.8158 |

Table 5: Comparisons on weeds dataset using Different Model.

A representation of our final results from our proposed method is shown in Fig. 7, and also for comparison we have shown results of R-FCN model in Fig 6. It can be clearly seen in Fig. 6 that our proposed method identifies more accurate objects than R-FCN.

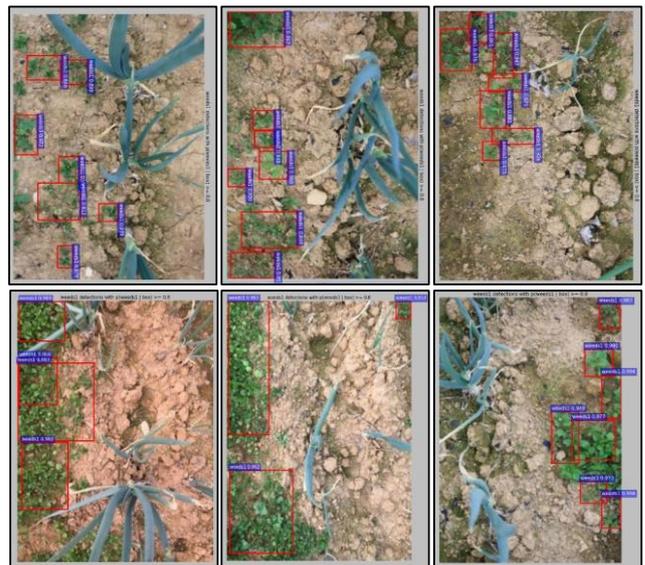

Figure 6: Sample results of the R-FCN.

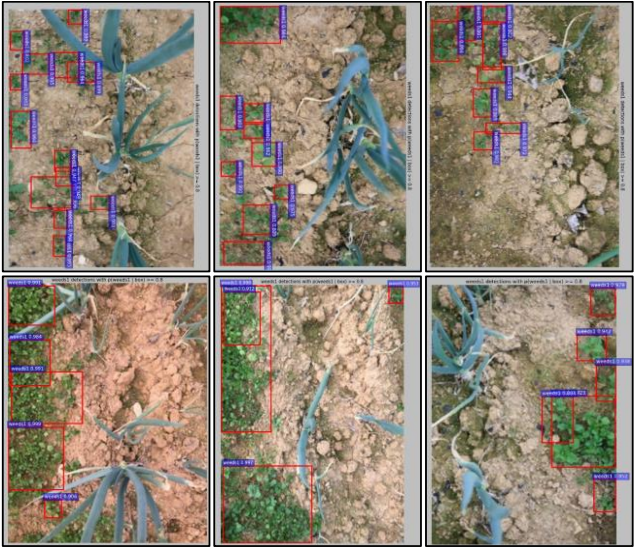

Figure 7: Sample results of the Proposed model.

## IV. Discussion

In this dissertation, we explored the conventional region based CNN models which was showed in previous state-of-art articles. We explored those models for building a strong object detector, so that we can obtain a better practical notion of their working principles. Moreover, we tested the most recent region based CNN models (Faster RNN and R-FCN) and by comparing these models we choose the optimal one for our proposal.

To understand the optimal model, we first proposed several alternative strategies to improve the performance of the ResNet. Among those strategies, one strategy where the ResNet contained a dropout layer after the first pooling layers outperformed in comparison to others. The reason behind the outstanding performance is that the proposed network opposed the overfitting for our comparatively small weeds dataset. Moreover, for finding the finest region based network we compared between the Faster RCNN and R-FCN by using our weeds detecting dataset and Figured out that R-FCN outpaces the Faster RCNN. For this reason, we chose the modified ResNet and R-FCN and suggested this network combination as the proposed architecture.

Scrutiny of Figure 6 and Figure 7 explicitly highlights that the proposed architecture can efficiently recognize almost all of the objects in the network. Moreover, an analysis of Figure 8 and Figure 9 supports our prediction about the recognition of objects as it distinctively identifies almost all the objects. This implies that our proposed network is capable of recognizing these entities due to local features exploration.

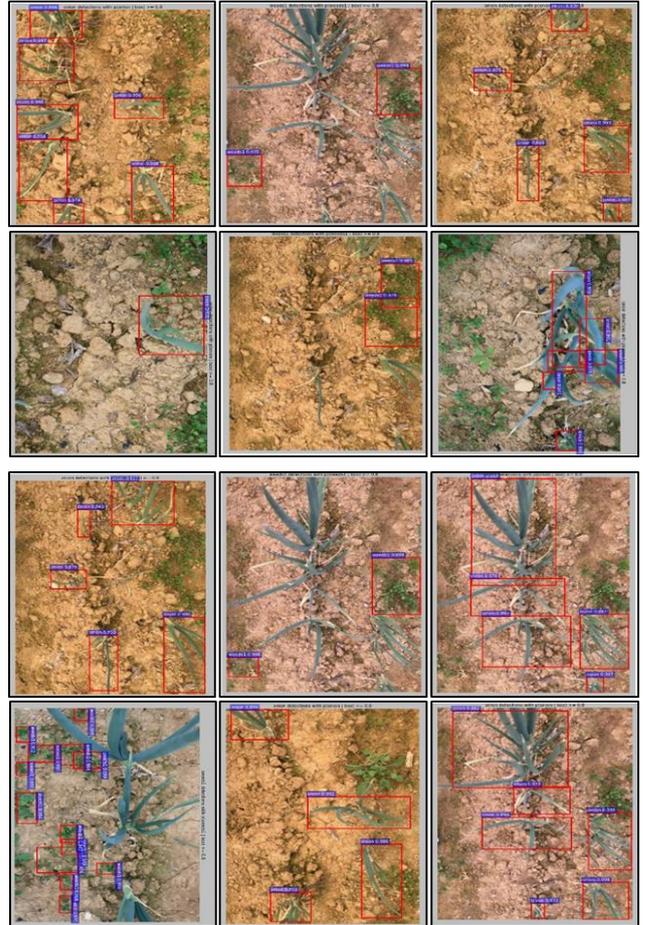

**Figure 8:** Examples of the Proposed architecture.

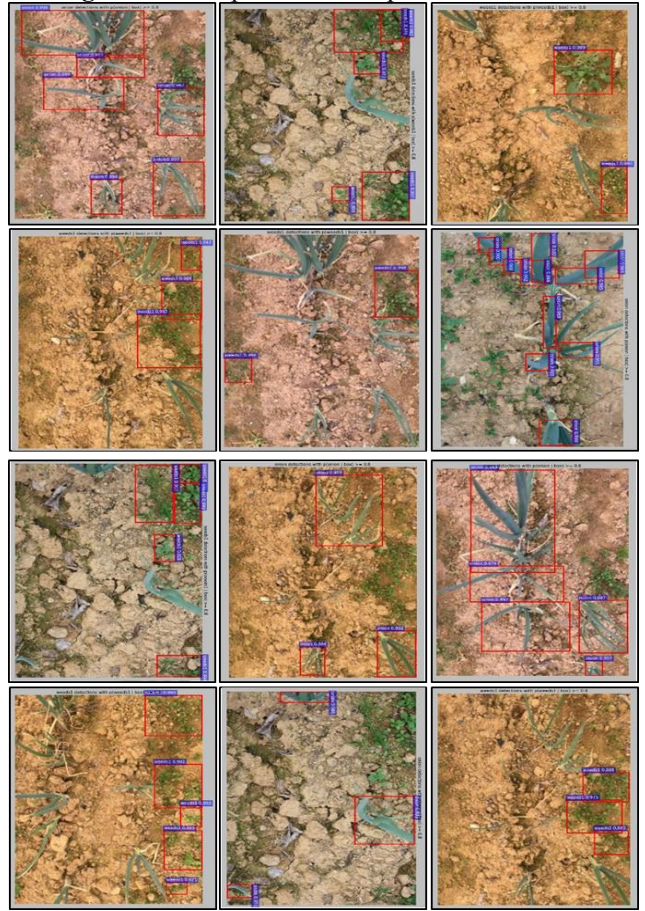

**Figure 9** Curated examples of the Proposed model.

## V. Conclusion

This paper has introduced a new alternative model called Region based Fully Convolutional Network. It can also be classified as R CNN series, which makes full use of the current classification of the best ResNet network, both in accuracy and speed when compared Faster R-CNN and has greatly improved. R-FCN uses position-sensitive score map to extract the localization translation from the network structure, and can make full use of ResNet's powerful classification capability by having a dropout layer for making our proposed system more stable with respect to new data for weeds detection in onion farm. By further increasing the learning rates, re-moving ReLu, re-moving batch normalization, and applying other modifications afforded by Dropout, we reached the previous state of art algorithm that created overfitting in our network. Furthermore, by combining multiple models trained with Dropout, it performs better than the best-known system on our weeds dataset. We consider that it is an important contribution to weeds detection. The proposed network is simple but more accurate and effective framework for farm land weeds detection. We believe this model can lead to even more insightful discoveries if it is studied on larger datasets.

## Acknowledgment

This work was carried out with the support of " Development of real-time diagnosis analysis technology for the major disease and insects of tomato using neural networks (Project No. PJ0120642016)", BK 21 Plus and Intelligent Robot Research Center, Chonbuk National University, Republic of Korea.